\pdfoutput=1
\documentclass[11pt, letterpaper, logo, onecolumn, copyright]{template_from_google}
\usepackage[T1]{fontenc}
\usepackage[utf8]{inputenc}
\usepackage{CJKutf8} % For Chinese characters

\usepackage{amsmath}
\usepackage{amsfonts}
\usepackage{amssymb}
\usepackage{amsthm}
\usepackage{mathtools}
\usepackage{bm} % For bold math symbols like \bm
\usepackage{nicefrac} % For typesetting compact fractions

\usepackage{booktabs} % For professional quality tables
\usepackage{enumitem} % For list customization
\usepackage{fancyhdr}
\usepackage{multirow} % For multi-row cells in tables
\usepackage{tabularx}
\usepackage{multicol}
\usepackage{array}
\usepackage{longtable}
\usepackage{lipsum} % For placeholder text
\usepackage{wrapfig} % To wrap text around figures
\usepackage{tocloft} % To control the Table of Contents
\usepackage{fancybox}
\usepackage{authblk} % For author affiliations
\usepackage{pifont} % For \cmark/\xmark symbols
\usepackage{microtype}

\usepackage{algorithm}
\usepackage[noend]{algpseudocode}
\usepackage{algpseudocode}

\usepackage{graphicx}
\usepackage{caption}
\usepackage{subcaption}
\usepackage{capt-of} % For using \captionof
\usepackage[inkscapeformat=png]{svg}

\usepackage{listings, listings-rust}
\usepackage{listingsutf8}

\usepackage[dvipsnames]{xcolor}
\usepackage[most, breakable, skins]{tcolorbox}
\tcbuselibrary{listings,skins,breakable}

\usepackage{fontawesome5}
\usepackage{xspace} % For intelligent spacing after macros

\usepackage[authoryear, sort&compress, round]{natbib}

\usepackage[dvipsnames]{xcolor} % Make sure xcolor is loaded

\usepackage{hyperref}

\definecolor{darkblue}{rgb}{0.0, 0.0, 0.6}
\definecolor{darkred}{rgb}{0.7, 0.0, 0.0}
\hypersetup{
  pdffitwindow=true,
  pdfstartview={FitH},
  pdfnewwindow=true,
  colorlinks,
  linktocpage=true,
  linkcolor=darkred,
  urlcolor=darkblue,
  citecolor=darkblue
}

\usepackage{amsmath,amsfonts,bm}

\def\eqref#1{equation~\ref{#1}}
\def\1{\bm{1}}

\DeclareMathAlphabet{\mathsfit}{\encodingdefault}{\sfdefault}{m}{sl}
\SetMathAlphabet{\mathsfit}{bold}{\encodingdefault}{\sfdefault}{bx}{n}

\newcommand{\pdata}{p_{\rm{data}}}
\newcommand{\E}{\mathbb{E}}

\newcommand{\KL}{D_{\mathrm{KL}}}

\usepackage[export]{adjustbox}
\usepackage[capitalise]{cleveref}
\crefname{equation}{Eq.}{Eqs.}

\usepackage{multirow}
\usepackage{caption}

\usepackage{paralist}

\providecommand{\E}{\mathbb{E}}
\providecommand{\KL}{D_{\mathrm{KL}}}

\providecommand{\pdata}{p_{\text{data}}}

\providecommand{\cmark}{\ding{51}}
\providecommand{\xmark}{\ding{55}}

\theoremstyle{plain}
\newtheorem{theorem}{Theorem}[section]

\newtheorem{corollary}[theorem]{Corollary}
\theoremstyle{definition}

\theoremstyle{remark}
\newtheorem{remark}[theorem]{Remark}

\makeatletter
\renewcommand\paragraph{\@startsection{paragraph}{4}{\z@}%
            {-2.5ex\@plus -1ex \@minus -.25ex}%
            {1.25ex \@plus .25ex}%
            {\itshape\normalsize\bfseries}}
\makeatother

\let\cite\citep

\title{Learning Discrete Autoregressive Priors with
Wasserstein Gradient Flow}
\newcommand{\shorttitle}{Learning Discrete Autoregressive Priors with
Wasserstein Gradient Flow}
\renewcommand{\headrulewidth}{0.4pt}

\reportnumber{} % Leave blank if n/a

\renewcommand{\today}{}

\author[1]{Bowen Zheng}
\author[2]{Yihong Luo}
\author[1]{Tianyang Hu}

\affil[1]{The Chinese University of Hong Kong, Shenzhen}
\affil[2]{Hong Kong University of Science and Technology}

\begin{abstract}
Discrete image tokenizers are commonly trained in two stages: first for reconstruction, and then with a prior model fitted to the frozen token sequences. This decoupling leaves the tokenizer unaware of the model that will later generate its tokens. As a result, the learned tokens may preserve image information well but still be difficult for an autoregressive (AR) prior to predict from left to right. We analyze this mismatch using Tripartite Variational Consistency (TVC), which decomposes latent-variable learning into three consistency conditions: conditional-likelihood consistency, prior consistency, and posterior consistency. TVC shows that two-stage training preserves the reconstruction side but leaves prior consistency outside the tokenizer objective: the overall token distribution is fixed before the AR prior participates in training. Motivated by this view, we add a distribution-level prior-matching signal during tokenizer training, while keeping the reconstruction objective unchanged. We optimize this signal with a Wasserstein-gradient-flow update. For hard categorical tokens, the update reduces to a token-level contrast between an auxiliary AR model that tracks the tokenizer's current token distribution and the target AR prior. It requires only forward passes through the two AR models and does not backpropagate through either of them. The resulting tokenizer, \textbf{wAR-Tok}, reduces AR loss and improves generation FID on CIFAR-10 and ImageNet at comparable reconstruction quality.
\end{abstract}

\begin{document}

\maketitle

\section{Introduction}\label{sec:1}
Latent generative models first compress an image into a shorter representation, and then learn a generative model over that representation~\citep{kingma2014vae}. This separation is used in Latent Diffusion Models~\citep{rombach2022ldm}, Masked Image Modeling~\citep{he2022mae}, and discrete image tokenizers. A common recipe is two-stage training~\citep{esser2021taming}: the tokenizer is first trained to reconstruct images, and a prior is then trained on the frozen token sequences. This recipe makes the tokenizer good at reconstruction, but it does not ask whether the resulting tokens are easy for the second-stage prior to model. In 1D tokenization~\citep{titok}, this issue becomes visible with autoregressive (AR) priors~\citep{sun2024llamagen}: the same tokens can reconstruct images well but remain difficult to predict from left to right.

Recent AR-alignment methods try to make the learned tokens more sequential. Tail dropout~\citep{soundstream,flextok} trains the decoder to reconstruct from prefixes, encouraging earlier tokens to carry useful information. Diffusion or flow-matching decoders~\citep{lipman2022flowmatching} introduce an ordered refinement process. SelfTok~\citep{selftok} ties the prefix length to the diffusion timestep. These methods can improve the ordering of information in the token sequence, but the supervision still comes through the decoder. The AR prior itself is not used to train the tokenizer, so its modeling difficulty is only addressed indirectly.

We use Tripartite Variational Consistency (TVC, \S\ref{sec:3}) to make this missing link explicit. TVC views latent-variable learning as the consistency of three parts: the conditional likelihood, the prior, and the posterior. This perspective separates two different failure modes. ELBO-style training forces each input's latent distribution to look like the prior, which can remove the information that the latent should carry. Two-stage training avoids this collapse by training the tokenizer only for reconstruction, but then leaves the entire burden of prior matching to the second stage. In other words, the tokenizer fixes a latent distribution before the AR prior ever participates in training.

This analysis points to a more direct objective. The tokenizer should still be trained for reconstruction, but it should also receive a signal that makes its overall token distribution closer to what the AR prior can model. We call this term Distributional Prior Divergence (DPD). Unlike the instance-wise constraint in ELBO, DPD only matches the aggregate distribution of tokens. Unlike standard two-stage training, it sends this prior-matching signal back to the tokenizer. Existing aggregate-matching methods such as WAE-MMD~\citep{tolstikhin2018wae}, WAE-GAN/AAE~\citep{makhzani2015adversarial}, and InfoVAE~\citep{zhao2017infovae} pursue a related goal in continuous latent spaces, but they do not directly apply to hard discrete tokenizers. Directly backpropagating the AR loss through the prior~\citep{wang2024larp} is also problematic, since it can favor trivial sequences that are easy to predict but less useful as image representations.

The remaining question is how to optimize this distribution-level mismatch for discrete tokens. We do this with a Wasserstein-gradient-flow update. Intuitively, the update compares two token-level predictions: one from a proxy model that tracks the tokenizer's current token distribution, and one from the target AR prior. If the proxy assigns higher probability to a token than the target prior does, the tokenizer is pushed away from that choice; if the target prior assigns higher probability, the tokenizer is pulled toward it. This gives an encoder-side prior-alignment signal using only forward passes through the proxy and the prior. No backpropagation through either model is required. We call the method \textbf{WGF-DPD}, and the tokenizer trained with it \textbf{wAR-Tok}.

On CIFAR-10, under matched reconstruction quality (rFID $\approx 5.0$), wAR-Tok improves gFID from $17.60$ to $14.50$ relative to the strongest baseline and reduces AR generation loss from $0.78$ to $0.42$. Under matched training iterations, it reaches gFID $11.30$ and gLoss $0.34$. On ImageNet $256{\times}256$, using the same encoder--decoder architecture and AR prior as the baseline, wAR-Tok reduces gFID from $6.32$ to $5.42$ and evaluation AR loss from $7.89$ to $7.07$, with rFID changing from $2.05$ to $2.35$.

\section{Preliminaries}\label{sec:2}
We consider the problem of learning a generative model for high-dimensional data $x \in \mathcal{X}$ drawn from an unknown data distribution $\pdata(x)$.
Latent Variable Models (LVMs) address this by introducing an unobserved latent variable $z \in \mathcal{Z}$, positing that the complex data distribution arises from a structured joint distribution $p_\theta(x, z)$.
The marginal likelihood of the model is obtained by integrating over the latent space:
\begin{equation*}
    p_{\theta}(x) = \int p_{\theta}(x \mid z)\, p_{\theta}(z) \, \mathrm{d}z,
\end{equation*}
where $p_\theta(z)$ is the prior and $p_\theta(x \mid z)$ is the conditional likelihood (or decoder).
The objective of an LVM is to optimize $\theta$ such that the model marginal $p_\theta(x)$ approximates the empirical data distribution $\pdata(x)$ as closely as possible.

\subsection{Evidence Lower Bound and Two-Stage Training}

Given a dataset $X$, the Evidence Lower Bound (ELBO)~\citep{kingma2014vae,rezende2014stochastic} is
\begin{align*}
\sum_{x \in X} \log p_{\theta}(x) &\ge \sum_{x}\Big(\E_{z\sim q_{\phi}(z\mid x)}\big[\log p_{\theta}(x\mid z)\big] \nonumber  - \KL\bigl(q_{\phi}(z\mid x)\,\|\, p_{\theta}(z)\bigr)\Big).
\end{align*}

Let $
q_\phi(z)=\int \pdata(x)\,q_\phi(z\mid x)\,\mathrm{d}x
$ denote the aggregate posterior. Expanding the KL term in the ELBO yields
\begin{align*}
&\E_{x\sim \pdata}\!\left[\KL\!\left(q_\phi(z\mid x)\,\|\,p_\theta(z)\right)\right]
= H_q\!\left(Z\right) - H_q\!\left(Z\mid X\right) + \KL\!\left(q_\phi(z)\,\|\,p_\theta(z)\right).
\end{align*}

Under deterministic coding, $H_q(Z\mid X)=0$, and the objective can be further written as
\begin{align*}
&\E_{z\sim q_{\phi}(z\mid x)}\big[\log p_{\theta}(x\mid z)\big]
- \big(H_q(Z)-H_q(Z\mid X)\big) 
- \KL\bigl(q_{\phi}(z)\,\|\,p_{\theta}(z)\bigr) \nonumber \\
&=
\underbrace{\E_{z\sim q_{\phi}(z\mid x)}\big[\log p_{\theta}(x\mid z)\big]}_{\text{reconstruction}}
+
\underbrace{\E_{z\sim q_{\phi}(z\mid x)}\big[\log p_{\theta}(z)\big]}_{\text{prior mapping}}.
\end{align*}

Modern discrete generative models can be viewed as a \textit{two-stage} training under this objective~\citep{oord2017vqvae,esser2021taming}:
the first stage optimizes the reconstruction term, while the second stage fits $q_{\phi}(z)$.

\subsection{Autoregressive Modeling}
Standard AR approaches~\citep{van2016pixelcnn} model the aggregate posterior $q_\phi(z)$ by maximizing the log-likelihood of the latent codes $z = (z_1, \dots, z_n)$ under a parametric prior $p_\theta(z)$. This is typically optimized via cross-entropy with teacher forcing~\citep{williams1989learning}:
\begin{align*}
\mathcal{L}_{\text{AR}}(\theta)
&= - \E_{z \sim q_\phi(z)} \left[ \sum_{t=1}^n \log p_\theta(z_t \mid z_{<t}) \right],
\end{align*}
where $z_{<t} \triangleq (z_1, \dots, z_{t-1})$ denotes the history of previous tokens.

\subsection{Existing AR-Alignment Techniques}
\label{sec:analysis}
Three representative techniques recur across recent work: \textbf{Tail Dropout}, \textbf{Diffusion Decoder}, and \textbf{Tied Timesteps}. As a controlled baseline we also consider \textbf{1D Tokenization}, which removes spatial biases but does not explicitly target AR alignment.

\noindent\textbf{1D Tokenization}~\citep{titok} feeds learnable holistic-token queries to the encoder rather than pixel embeddings, yielding latent representations decoupled from explicit 2D structure.

\noindent\textbf{Tail Dropout}, first explored in audio codecs~\citep{soundstream} and adapted to vision~\citep{flextok}, randomly discards a suffix of the token sequence during training and requires the decoder to reconstruct from any prefix:
\begin{equation*}
\mathcal{L}_{\text{tail}} = \sum_{k=0}^{n} \log p_\theta(x \mid z_0, z_1, \ldots, z_k).
\end{equation*}

\noindent\textbf{Implicit AR Prior by Diffusion Decoder}~\citep{lipman2022flowmatching} replaces the deterministic decoder with a flow-matching one, introducing sequential structure indirectly through the denoising trajectory.

\noindent\textbf{Tied Timesteps}~\citep{selftok} explicitly couples the tail-dropout ratio with the diffusion timestep, binding the discrete prefix length to a noise level. We discuss these methods further in Sec.~\ref{sec:6}.

\section{Tripartite Variational Consistency: A Framework}\label{sec:3}
\label{sec:tvc_framework}

TVC is an analytical framework for latent-variable objectives. The decomposition follows from Bayes' rule on a data-anchored joint, and we treat it as a tool for organizing the design space rather than as a theoretical contribution. Section~\ref{sec:tvc-design-space} uses it to make the design space explicit, and the WGF-DPD choice in \S\ref{sec:4} occupies one specific corner of that space.

Latent-variable methods introduce an auxiliary variable $z$, positing that high-dimensional observations $x$ admit a structured representation in a latent space. To formalize this, we consider a \textit{data-anchored variational joint distribution}
$q_\phi(x,z)\triangleq \pdata(x)\,q_\phi(z\mid x)$.
The corresponding latent marginal
$q_\phi(z)=\int \pdata(x)\,q_\phi(z\mid x)\,\mathrm{d}x$
represents the aggregate posterior induced by the encoder. This construction reframes learning as alignment between the model joint $p_\theta(x,z)$ and the data-anchored joint $q_\phi(x,z)$.

Both prevailing paradigms struggle with this alignment on the prior side. ELBO-based methods impose restrictive instance-wise constraints that often precipitate posterior collapse. Two-stage approaches decouple the optimization, frequently yielding latent spaces that are structurally hostile to the prior and that demand ad-hoc tricks or massive parameter scaling to compensate.

\textbf{Tripartite Variational Consistency (TVC)} leverages Bayes' rule to decompose the global generative objective into three coupled components: the conditional likelihood, the prior, and the posterior.

\begin{theorem}[Tripartite Variational Consistency]
\label{thm:consistency}
The log-density ratio between the model distribution $p_\theta(x)$ and the data distribution $\pdata(x)$ admits the following decomposition:
\begin{equation}
    \begin{aligned}
    \log p_\theta(x)
    &= \underbrace{\log \frac{p_\theta(x\mid z)}{q_\phi(x\mid z)}}_{\text{Likelihood Consistency}}
     + \underbrace{\log \frac{p_\theta(z)}{q_\phi(z)}}_{\text{Prior Consistency}} 
     \quad- \underbrace{\log \frac{p_\theta(z\mid x)}{q_\phi(z\mid x)}}_{\text{Posterior Consistency}}
     + \log \pdata(x)
    \end{aligned}
\end{equation}
\end{theorem}

Achieving perfect generation $p_\theta(x) = \pdata(x)$ is equivalent to satisfying these three consistency constraints simultaneously.
All proofs can be found in \textbf{Appendix~\ref{sec:proof}}.

We formulate TVC as a \textbf{consistency criterion} that aligns $p_\theta(x)$ with $\pdata(x)$ by enforcing constraints along the \textbf{three component edges} of the joint distributions: the Conditional Likelihood, the Prior, and the Posterior. The decomposition rests on a single move: applying Bayes' rule to a carefully constructed data-anchored joint distribution yields a transparent structural decomposition of the generative objective.

\subsection{Constraint Redundancy}

Theorem~\ref{thm:consistency} delineates three distinct consistency conditions, but the constraints are coupled. Enforcing consistency on specific subsets of these components is sufficient to guarantee global optimality.

\begin{corollary}[Constraint Redundancy]
\label{lem:redundancy}
The three consistency conditions in Theorem~\ref{thm:consistency} exhibit mutual dependency, such that satisfying specific pairs implies the third, guaranteeing $p_\theta(x) = \pdata(x)$.
\end{corollary}
This redundancy manifests in three distinct forms: 1) \textbf{Likelihood--Prior Matching}: Given the alignment of likelihoods $p_\theta(x\mid z) = q_\phi(x\mid z)$ and priors $p_\theta(z) = q_\phi(z)$, Posterior Consistency follows automatically. 2) \textbf{Likelihood--Posterior Matching}: Similarly, the conjunction of Likelihood Consistency and Posterior Consistency $p_\theta(z\mid x) = q_\phi(z\mid x)$ necessitates the alignment of the priors. 3) \textbf{Posterior--Prior Matching}: The implication is more restrictive for the final pair. The simultaneous consistency of the posterior and prior distributions guarantees Likelihood Consistency provided that the family of variational distributions $\{q_\phi(x\mid z)\}_z$ satisfies the completeness condition.

\subsection{TVC Unifies ELBO and Two-Stage Training}\label{sec:tvc-instances}

Several established generative modeling paradigms can be rigorously interpreted as specific instances of the TVC formalism, utilizing the redundancy properties established in~\ref{lem:redundancy}.

\noindent\textbf{ELBO as an Instance of TVC.} The ELBO objective is a specific instantiation of the TVC formalism that leverages the redundancy property (Corollary~\ref{lem:redundancy}, Case 1). By explicitly optimizing the Likelihood and Prior consistency terms with KL divergence, it implicitly enforces Posterior consistency.

\noindent\textbf{ELBO Variants as an Instance of TVC.} Many ELBO variants that mitigate posterior collapse, including InfoVAE~\citep{zhao2017infovae}, AAE~\citep{makhzani2015adversarial}, $\beta$-VAE~\citep{Higgins2016betaVAELB}, and WAE~\citep{tolstikhin2018wae}, can be viewed as alternative relaxations of the ELBO prior-matching term that still target the \textbf{Likelihood--Prior Matching} redundancy (Corollary~\ref{lem:redundancy}, Case~1), and hence the same marginal goal $p_\theta(x)\approx \pdata(x)$. See Appendix~\ref{Proof:ELBOVariants_TVC} for a unified formulation and derivation from TVC.

\noindent\textbf{Two-Stage Models as an Instance of TVC.} Two-stage discrete generative models, such as VQ-VAE~\citep{oord2017vqvae} and VQGAN~\citep{esser2021taming}, instantiate the TVC formalism by sequentially satisfying the constraints of \textbf{Likelihood--Prior Matching} (Corollary~\ref{lem:redundancy}, Case 1). Decoupling optimization into a reconstruction phase and a prior alignment phase achieves global distributional identity without maximizing a joint bound.

A practical benefit of the formalism is that it decouples the choice of objective, optimization procedure, and distributional discrepancy. InfoVAE-style objectives fit naturally here: as long as the resulting training drives the model toward the TVC consistency regime, the specific weighting hyperparameters can be viewed as empirical knobs for optimization rather than quantities requiring method-specific derivations. This provides a clear target for exploring objectives and divergences that are more favorable for optimization while preserving the same marginal goal.

\subsection{TVC as a Design Space}\label{sec:tvc-design-space}

A central practical benefit of TVC is that it decouples the choice of objective along three orthogonal axes. \textbf{(A1) Consistency-pair decoupling.} Most existing methods target Likelihood--Prior. Corollary~\ref{lem:redundancy} identifies two additional valid pairings. \textbf{(A2) Divergence decoupling.} TVC is divergence-agnostic. Each consistency term may use a different divergence (KL, MMD, Wasserstein, $f$-divergence, $\dots$). \textbf{(A3) Optimization decoupling.} Different consistency terms admit different schemes (reparameterization, adversarial, score matching, Wasserstein gradient flow, $\dots$).

WGF-DPD (\S\ref{sec:4}) sits at axis A2 (KL on the prior) and axis A3 (Wasserstein gradient flow). To check that the axes generate objectives that are not just reparameterizations of existing ones, we sketch a separate construction below.

\begin{remark}[A novel objective from TVC]\label{rem:new-objective}
\textbf{(A1)} By Case~2, enforcing exact Likelihood and Posterior Consistency yields $p_\theta(x)=\pdata(x)$. \textbf{(A2)} Assign KL to the likelihood term and MMD to the posterior term independently:
$\mathcal{L}=\E_{z\sim q_\phi}\!\big[\KL(q_\phi(x\mid z)\|p_\theta(x\mid z))\big]+\lambda\,\E_{x\sim\pdata}\!\big[\mathrm{MMD}^2(q_\phi(z\mid x),p_\theta(z\mid x))\big]$. \textbf{(A3)} Replace the MMD gradient with a normalized-kernel \emph{drifting} field $V(z\mid x)$ such that matching $q_\phi(z\mid x)=p_\theta(z\mid x)$ implies zero drift, and replace the KL likelihood term with the standard reconstruction surrogate. The result is
$\mathcal{L}_{\text{new}}=\E_{x,z\sim q_\phi}[-\log p_\theta(x\mid z)]+\lambda\,\E_{x,z\sim q_\phi}\!\big[\big\|z-\mathrm{sg}[z+V(z\mid x)]\big\|^2\big]$,
which lies outside the standard ELBO and two-stage menu. We do not pursue this objective empirically. The construction is included only to evidence that the three axes are productive.
\end{remark}

\section{TVC for Autoregressive Prior}\label{sec:4}

\subsection{Suboptimal Representations of ELBO and Two-Stage Training}\label{sec:why-discrete-fails}
The limitations of both paradigms stem from their treatment of Prior Consistency.
Under TVC, the ELBO can be interpreted as enforcing prior consistency through a strict \textbf{instance-wise} constraint. Specifically, satisfying $q_\phi(z\mid x)=p_\theta(z)$ for all $x \sim \pdata$ implies the alignment of the marginal distributions:
\(
q_\phi(z) = \int \pdata(x)\, q_\phi(z\mid x)\, \mathrm{d}x = p_\theta(z).\)
This condition is restrictive: strictly equating the conditional posterior to the prior implies that $z$ is statistically independent of $x$, causing the mutual information to vanish:
\(
 I(X;Z) = H_q(Z) - H_q(Z\mid X) = 0.
\)
This is the phenomenon of posterior collapse~\citep{bowman2016generating,Lucas2019UnderstandingPC}. Instance-wise alignment is a \textit{sufficient} but not \textit{necessary} condition for generative modeling. Optimizing the data log-likelihood requires only that the \textit{aggregate} posterior $q_\phi(z)$ matches the prior $p_\theta(z)$, which lets the encoder retain dependence on the input.

Two-stage methods take the opposite path. They first optimize the encoder $q_\phi(z\mid x)$ solely for reconstruction fidelity, leaving the aggregate posterior $q_\phi(z)$ unconstrained. The entire alignment burden then falls on the second stage. If the induced latent structure is statistically misaligned with the inductive bias of the AR prior (lacking causal ordering or containing high-frequency artifacts), fitting $p_\theta(z)$ becomes inefficient, and excessive parameter scaling is often needed to compensate.

\subsection{Easing Distributional Prior Divergence from TVC}
TVC suggests a different formulation: global prior consistency requires only that the \textit{marginal} distributions align. We retain the likelihood constraint (reconstruction) and relax the prior-matching term from a conditional to a distributional prior divergence (DPD):
\begin{equation}\label{eq:dpd}
\mathcal{L}_{DPD}= -\E_{x\sim p_{data}}\Big[\E_{z\sim q_\phi(z\mid x)} \big[\log p_\theta(x\mid z)\big]\Big] + \KL\!\left(q_\phi(z)\,\|\,p_\theta(z)\right).
\end{equation}

This design preserves the established reconstruction objective while resolving the prior consistency limitations of both ELBO and two-stage training, and is conceptually related to distribution-level aggregate-posterior matching objectives explored in adversarial and Wasserstein autoencoders~\citep{tolstikhin2018wae} and information-regularized VAEs~\citep{zhao2017infovae}.

To practically optimize $\mathcal{L}_{\text{DPD}}$, we need gradients with respect to both $\theta$ and $\phi$. While $\nabla_\theta \mathcal{L}_{\text{DPD}}$ is accessible via standard backpropagation, $\nabla_\phi \mathcal{L}_{\text{DPD}}$ is hard, particularly when $q_\phi(z\mid x)$ takes a complex or non-reparameterizable form. The existing distribution-level aggregate-posterior matching methods (WAE-MMD, WAE-GAN/AAE, InfoVAE) all break down on hard discrete quantizers, in distinct ways summarized in Appendix~\ref{app:wae-comparison}.

\subsection{Effective Optimization of DPD}

\textbf{WGF-DPD} is a particle-based optimization method grounded in the theory of Wasserstein Gradient Flows~\citep{doi:10.1137/S0036141096303359,Otto31012001,ambrosio2005gradient,santambrogio2015optimal}. Rather than differentiating with respect to $\phi$ directly, we model the optimization as the evolution of the probability measure $q$ along the steepest descent direction of the energy functional $\mathcal{L}_{\text{DPD}}$ within the Wasserstein-2 space.

Decomposing the parameter update by chain rule through the particle positions, the gradient is approximated as
\begin{equation*}
\nabla_\phi \mathcal{L}_{\text{DPD}} \approx \E_{z \sim q_\phi} \left[ \left(\frac{\partial z}{\partial \phi}\right)^\top \mathbf{v}(z) \right],
\end{equation*}
where $\mathbf{v}(z)$ denotes the instantaneous optimal transport velocity field. In the Wasserstein-2 geometry, this velocity is defined by the negative gradient of the first variation of the functional:
\begin{equation*}
\mathbf{v}(z) \triangleq -\nabla_z \left( \frac{\delta \mathcal{L}_{\text{DPD}}}{\delta q} \right)(z).
\end{equation*}

Background on Wasserstein Gradient Flow is in Appendix~\ref{chap:AppendixB}.

\subsection{Gradient Flow Derivation and Tractable Approximation}
The gradient of $\mathcal{L}_{\text{DPD}}$ with respect to a latent sample $z$ comprises two components. The reconstruction term acts as a standard potential field. The prior matching term is a functional of the marginal distribution $q(z)$ and is optimized via Wasserstein Gradient Flow.

Applying the first variation to the KL divergence term $\mathcal{F}[q] = \KL(q\|p)$, we obtain the velocity field
\begin{align*}
\nabla_z \frac{\delta \mathcal{F}}{\delta q(z)}
&= \nabla_z \left( \log \frac{q(z)}{p_\theta(z)} + 1 \right) = \nabla_z \log q(z) - \nabla_z \log p_\theta(z).
\end{align*}

Combining with the reconstruction gradient, the total particle update direction becomes
\begin{equation}\label{eq:wgf-cont}
\nabla_z \mathcal{L}_{\text{DPD}} = \underbrace{-\nabla_z \log p_\theta(x\mid z)}_{\text{Reconstruction Score}} + \underbrace{\nabla_z \log q(z) - \nabla_z \log p_\theta(z)}_{\text{Prior Matching Score}}.
\end{equation}
The exact aggregate score $\nabla_z \log q_\phi(z)$ is intractable: computing it requires integrating the conditional encoder over the entire data distribution, $q_\phi(z) = \E_{x\sim \pdata}[q_\phi(z\mid x)]$, and batch-level estimates would introduce high variance and bias.

To make the functional gradient computable, we introduce a parametric auxiliary model $q_\psi(z)$ to approximate the current aggregate posterior $q_\phi(z)$. Substituting this proxy into the gradient equation yields a tractable update:
\begin{equation*}
\nabla_z \mathcal{L}_{\text{DPD}} \approx \underbrace{-\nabla_z \log p_\theta(x\mid z)}_{\text{Reconstruction Score}} + \underbrace{\nabla_z \log q_\psi(z) - \nabla_z \log p_\theta(z)}_{\text{Prior Matching Score}}.
\end{equation*}
Under standard Gaussian assumptions for the likelihood, the first term corresponds to the gradient of the Mean Squared Error (MSE). The latter terms are simply the score functions of the aggregate posterior proxy $q_\psi$ and the prior $p_\theta$. Optimization therefore requires only access to the \textit{scores} (gradients) of these distributions, not their normalizing constants.

\subsection{Score Estimation in Discrete Latent Spaces}
In discrete generative model, the $p_\theta(z), q_\psi(z)$ are categorical distributions. Assume the classes number are $K$, the one-hot based form can be written as
\begin{equation*}
P(z \mid \mathbf{p})=\prod_{k=1}^K p_k^{z_k},\quad z =  (\delta_{1i}, \ldots, \delta_{Ki})
\end{equation*}

where $p_k$ is the normalized probability of each class. Treating $z$ as continuous (e.g., via continuous relaxations such as the Gumbel-Softmax/Concrete distribution~\citep{jang2017gumbelsoftmax,maddison2017concrete}) gives
\begin{equation*}
\nabla_z \log P(z\mid \mathbf{p}) = \frac{\partial \sum_{k=1}^K z_k\log p_k }{\partial z} = \log \mathbf{p}
\end{equation*}

Substituting the derived categorical scores into the gradient flow equation, the particle-level gradient for a discrete latent code $z$ becomes
\begin{equation}\label{eq:wgf-discrete}
\nabla_z \mathcal{L}_{\text{DPD}} \approx -\nabla_z \log p_\theta(x\mid z) + \log \mathbf{Q}_\psi - \log \mathbf{P}_\theta.
\end{equation}

Here $\log \mathbf{P}_\theta$ and $\log \mathbf{Q}_\psi$ denote the \emph{per-token} log-probability vectors (over $K$ codebook entries) induced by the AR factorizations $p_\theta(z)=\prod_{t=1}^n p_\theta(z_t\mid z_{<t})$ and $q_\psi(z)=\prod_{t=1}^n q_\psi(z_t\mid z_{<t})$.

To propagate this signal to the encoder parameters $\phi$, we must traverse the non-differentiable quantization step \(z = F_{\text{quant}}(h),\)
where $z \in \{0,1\}^K$ is the one-hot representation of a discrete token and $h \in \Delta^{K-1}$ is its associated probability vector on the simplex.
We obtain $h=\mathrm{softmax}(\ell)$ from normalized logits $\ell$, either naturally arising in Gumbel-based quantizers or computed from vector-quantization distances (e.g., $\ell_k \propto -\|u-c_k\|_2^2$ for a latent embedding $u$ and codebook entries $\{c_k\}_{k=1}^K$), and define the hard assignment by \(z = \mathrm{onehot}\!\left(\arg\max_k h_k\right).\)
The Straight-Through Estimator (STE)~\citep{bengio2013stochastic} treats the backward Jacobian as $\partial z/\partial h \approx I$, giving
\begin{equation*}
\begin{aligned}
\nabla_\phi \mathcal{L}_{\text{DPD}}
&= \E \left[ \left(\frac{\partial z}{\partial \phi}\right)^\top \nabla_z \mathcal{L}_{\text{DPD}} \right] 
\approx \E \left[ \left(\frac{\partial h}{\partial \phi}\right)^\top \underbrace{\frac{\partial z}{\partial h}}_{\approx I} \Big( \underbrace{-\nabla_z \log p_\theta(x\mid z)}_{\text{Recon. Grad}} + \underbrace{\log \mathbf{Q}_\psi - \log \mathbf{P}_\theta}_{\text{WGF Score Update}} \Big) \right].
\end{aligned}
\end{equation*}

Since $\log \mathbf{Q}_\psi$ and $\log \mathbf{P}_\theta$ are simply the log-probability vectors from the aggregate posterior proxy and the prior, this formulation does NOT require backpropagation through $p_\theta$ or $q_\psi$. Training is therefore highly efficient. It adds only two forward passes, which are significantly cheaper than full backpropagation.

\subsection{Algorithm and connection to score-matching distillation}\label{sec:algorithm}

\begin{algorithm}[t]
\caption{One training step of wAR-Tok (WGF-DPD).}
\label{alg:wgfdpd}
\begin{algorithmic}[1]
\State \textbf{Inputs:} batch $x$; encoder $E_\phi$; quantizer $F_{\text{quant}}$; decoder $D_\theta$; AR target prior $p_\theta(z)$; AR proxy $q_\psi(z)$.
\State $h\leftarrow\mathrm{softmax}(E_\phi(x))$;\quad $z\leftarrow F_{\text{quant}}(h)$ \Comment{Encode + quantize}
\State $\hat x\leftarrow D_\theta(z)$;\quad $\mathcal{L}_{\text{rec}}\leftarrow\ell_1+\ell_2+\text{LPIPS}$ ($+\,$GAN if used) \Comment{Reconstruction}
\State $\log\mathbf{P}_\theta\leftarrow\text{forward}(p_\theta,z)$;\quad $\log\mathbf{Q}_\psi\leftarrow\text{forward}(q_\psi,z)$ \Comment{Score forward (no backprop)}
\State $g_z\leftarrow-\nabla_z\log p_\theta(x\mid z)+(\log\mathbf{Q}_\psi-\log\mathbf{P}_\theta)$ \Comment{Particle gradient~\eqref{eq:wgf-discrete}}
\State $g_\phi\leftarrow$ backprop $g_z$ to $\phi$ via STE ($\partial z/\partial h\approx I$) \Comment{Encoder update}
\State $\mathcal{L}_{\text{AR}}\leftarrow$ cross-entropy of $p_\theta$ on $z$;\quad $\mathcal{L}_\psi\leftarrow$ cross-entropy of $q_\psi$ on $z$
\State Update $\{\phi,\theta,\psi\}$ jointly with AdamW \Comment{No alternating updates}
\end{algorithmic}
\end{algorithm}

Algorithm~\ref{alg:wgfdpd} summarizes one training step. All losses are optimized jointly with AdamW. The proxy $q_\psi$ is trained to track $q_\phi(z)$. A more expressive $q_\psi$ tracks $q_\phi$ more closely and improves the WGF-gradient estimate. We use a $\sim$17M-parameter AR proxy on CIFAR-10 and a $\sim$40M proxy on ImageNet-256.

\noindent\textbf{Relation to distribution-matching distillation.} The velocity field~\eqref{eq:wgf-cont} is the Wasserstein gradient of $\KL(q\|p)$, and it does not depend on whether $z$ is continuous or discrete. With diffusion-estimated scores in continuous spaces, the same score-difference form appears in Distribution Matching Distillation (DMD)~\citep{yin2024onestepdiffusiondistributionmatching} and Diff-Instruct~\citep{luo2024diffinstructuniversalapproachtransferring}. With AR factorizations in discrete categorical spaces, it reduces to the per-token log-probability vectors of~\eqref{eq:wgf-discrete}. WGF-DPD is therefore a discrete instantiation of the same gradient.

\subsection{Objective Decomposition and Behavior Analysis}
The prior-matching term of objective~\ref{eq:dpd} can be conceptually decomposed as
\begin{equation*}
\begin{aligned}
&\E_{z\sim q_\phi(z)}\big[\log q_\psi(z)-\log p_\theta(z)\big] = \KL\!\left(q_\phi(z)\,\|\,p_\theta(z)\right) - \KL\!\left(q_\phi(z)\,\|\,q_\psi(z)\right).
\end{aligned}
\end{equation*}
When $q_\psi$ matches $q_\phi$ perfectly, this surrogate objective recovers the exact prior-matching term $\KL\!\left(q_\phi(z)\,\|\,p_\theta(z)\right)$ (plus an entropy term). Even when $q_\psi$ does not match $q_\phi$ perfectly, optimizing this objective still pushes the true aggregate posterior away from $q_\psi$ and closer to our target $p_\theta$. The design therefore relies on two conditions. First, $q_\psi$ should track the aggregate posterior $q_\phi(z)$ closely. Second, $p_\theta(z)$ should provide a stronger autoregressive inductive bias than $q_\phi(z)$ currently possesses. Architecture and training settings for $q_\psi$ and $p_\theta$ are detailed in Appendix~\ref{chap:AppendixC}.

\begin{figure}[t]
\centering
\begin{subfigure}[t]{0.48\textwidth}
\centering
\includegraphics[width=\linewidth]{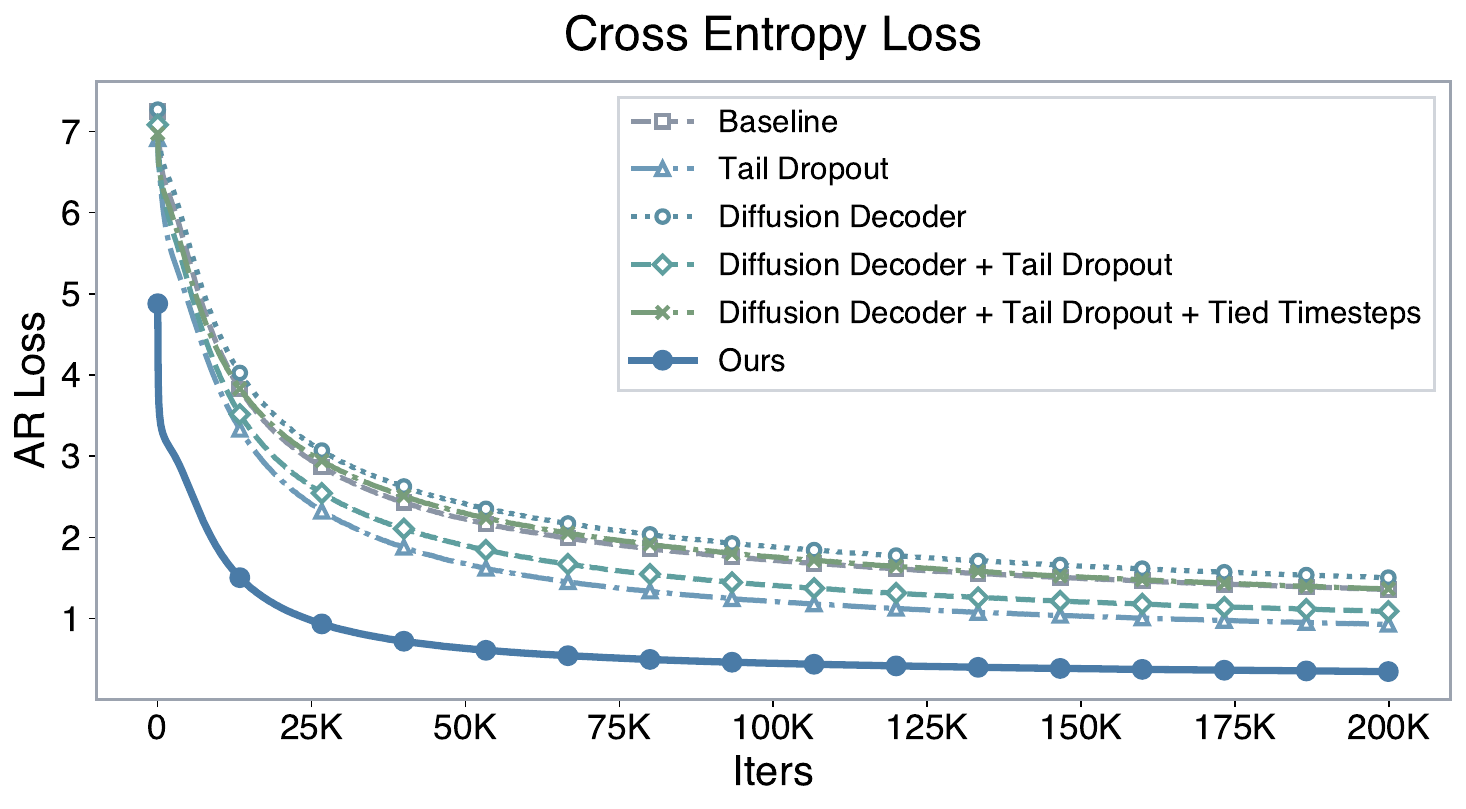}
\caption{Training loss curve.}
\label{fig:train_curves:loss}
\end{subfigure}\hfill
\begin{subfigure}[t]{0.48\textwidth}
\centering
\includegraphics[width=\linewidth]{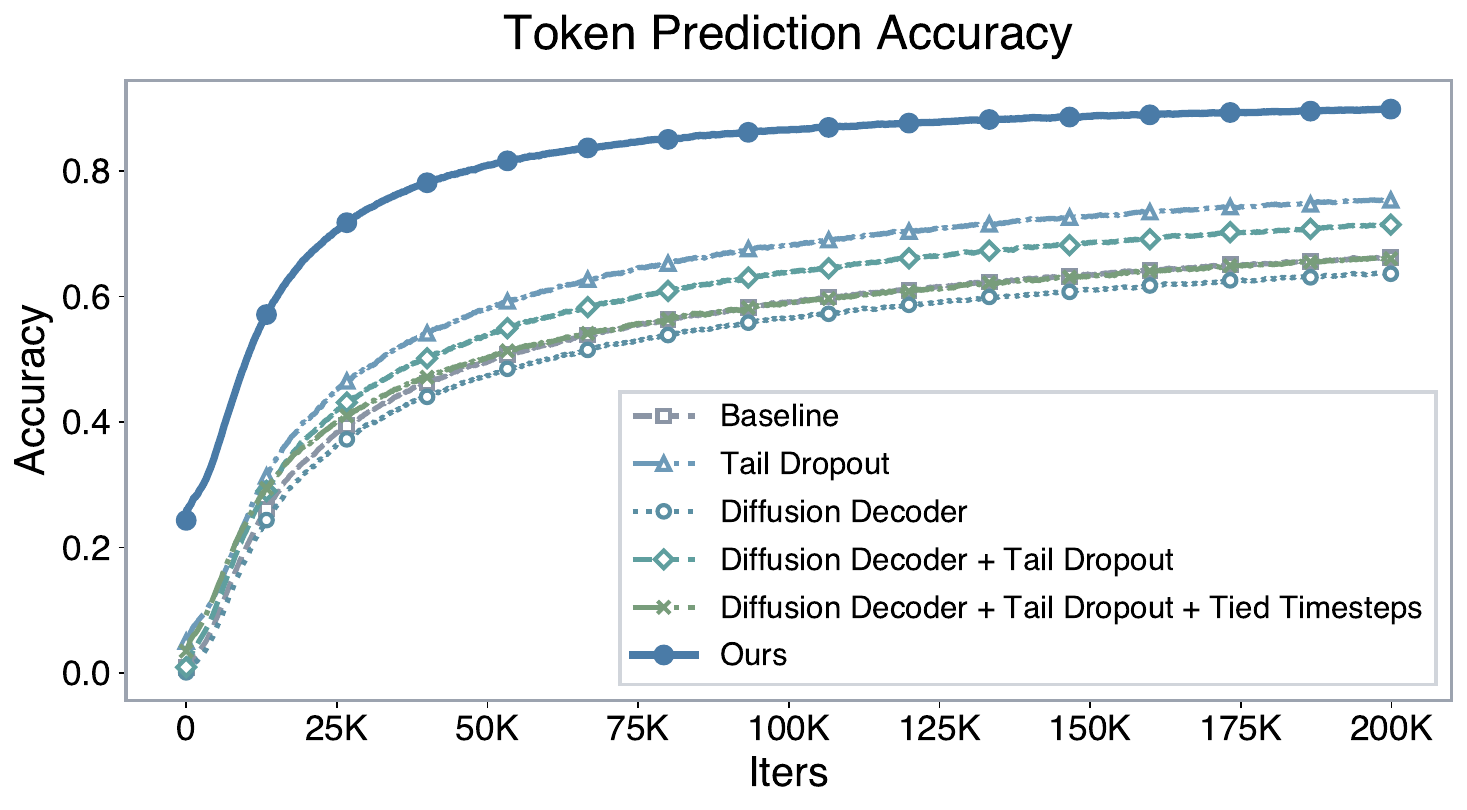}
\caption{Top-1 accuracy.}
\label{fig:train_curves:top1}
\end{subfigure}
\caption{AR training dynamics under matched training budgets on CIFAR-10. (a) Cross-entropy loss; (b) Teacher-forcing top-1 accuracy. wAR-Tok exhibits substantially lower loss and higher accuracy throughout training, indicating improved AR-alignment of the learned latent codes.}
\label{fig:train_curves}
\end{figure}

\section{Experiments}\label{sec:5}

We adopt 1D Tokenization as the foundational baseline, which provides a clean testbed without confounding spatial inductive biases. We evaluate \emph{Tail Dropout}, \emph{Diffusion Decoders}, and \emph{Tied Timesteps} against \emph{WGF-DPD}. Encoder, decoder, and quantizer architectures are fixed across all configurations, so only the generative-modeling paradigm varies. Adversarial training is excluded from all baselines on CIFAR-10 to eliminate confounding instability. Full architectural details and the evaluation protocol (matched-rFID and matched-iteration) are in Appendix~\ref{chap:AppendixC}. FID is computed with 50K samples following ADM~\citep{dhariwal2021diffusion}.

\subsection{CIFAR-10 Results}\label{sec:cifar}

Table~\ref{tab:compare_rfid} summarizes the matched-rFID protocol: wAR-Tok achieves the best gFID and the lowest AR generation loss (gLoss) while using fewer iterations than baselines. Under the matched-iteration protocol (Appendix~\ref{app:matched-iters}, Table~\ref{tab:compare_iters}), wAR-Tok maintains advantages in generation quality and AR loss while achieving comparable reconstruction. As training progresses, most baselines show limited gFID improvement or even degrade, whereas wAR-Tok continues to improve. Figure~\ref{fig:train_curves} shows the gap between wAR-Tok and the baselines becomes more pronounced as training progresses. Tied timesteps perform poorly: they require significantly longer training to reach comparable rFID and exhibit the worst gLoss, suggesting the denoising-trajectory ordering is misaligned with AR priors' inductive biases.

\begin{table}[t]
\centering
\caption{CIFAR-10, comparison under \textbf{comparable reconstruction quality (rFID $\approx 5.0$)}. With matched rFID, wAR-Tok attains superior generation quality (gFID) and AR loss.}
\label{tab:compare_rfid}
\footnotesize
\begin{adjustbox}{width=\textwidth}
\begin{tabular}{lcccccccc}
\toprule
\multirow{2}{*}{Representative Method} & \multirow{2}{*}{Iters $(\downarrow)$} & \multicolumn{4}{c}{Training Techniques} & \multirow{2}{*}{rFID} & \multirow{2}{*}{gFID $(\downarrow)$} & \multirow{2}{*}{gLoss $(\downarrow)$}\\
\cmidrule(lr){3-6}
& & FM Dec. & Tail Drop. & Tied Steps & WGF-DPD & & & \\
\midrule
TiTok~\citep{titok}                       & 50K  & \xmark & \xmark & \xmark & \xmark & 4.89 & 44.42 & 1.07\\
ImageFolder/One-D-Piece~\citep{imagefolder,onedpiece}  & 90K  & \xmark & \cmark & \xmark & \xmark & 4.94 & 17.60 & 0.78\\
FlowMo~\citep{sargent2025flowmodemodeseekingdiffusion}                     & 290K & \cmark & \xmark & \xmark & \xmark & 4.91 & 38.68 & 1.50\\
FlexTok~\citep{flextok}                   & 350K & \cmark & \cmark & \xmark & \xmark & 5.07 & 24.23 & 1.13\\
SelfTok~\citep{selftok}                   & 500K & \cmark & \cmark & \cmark & \xmark & 5.01 & 42.60 & 1.41\\
\midrule
\textbf{wAR-Tok (Ours)}                      & 100K & \xmark & \xmark & \xmark & \cmark & 4.99 & \textbf{14.50} & \textbf{0.42}\\
\bottomrule
\end{tabular}
\end{adjustbox}
\end{table}

Below wAR-Tok, tail dropout improves both gLoss and gFID, alleviating the AR hardness of the induced latents. Diffusion decoders, in contrast, converge substantially more slowly in Stage-I and do not improve downstream AR modeling in our setting: at similar rFID, diffusion-decoder variants exhibit higher gLoss than deterministic ones. While iterative diffusion sampling can partially compensate at inference and yield slightly better gFID than vanilla 1D tokenization, combining diffusion decoders with tail dropout makes the AR-hardness penalty more pronounced and yields worse gLoss/gFID than tail dropout alone.

\subsection{ImageNet 256$\times$256 Results}\label{sec:imagenet}

To check that the gains generalize beyond CIFAR-10, we run a controlled comparison on ImageNet $256{\times}256$. Both the baseline and wAR-Tok use a standard 1D-tokenizer architecture (encoder/decoder $\sim$86M parameters total), an AR target prior of $\sim$115M parameters trained for 300 epochs, and a $\sim$40M-parameter AR proxy $q_\psi$. Reconstruction losses are $\ell_1$ (weight 1.0), LPIPS (weight 1.0), and PatchGAN (weight 0.1). Both models use AdamW with weight decay $3\times10^{-2}$. The only training-recipe difference between baseline and wAR-Tok is the WGF-DPD term. Table~\ref{tab:imagenet256} reports the results: rFID is slightly higher under wAR-Tok (2.35 vs.\ 2.05 for the baseline), gFID drops from 6.32 to 5.42 (about 14\% relative improvement), and evaluation AR loss drops from 7.89 to 7.07. We exclude head-to-head comparisons against ImageFolder, FlowMo, FlexTok, and SelfTok because each relies on auxiliary techniques (DINO alignment, distillation, diffusion-decoder post-training) orthogonal to WGF-DPD. Convergence curves and full setup are in Appendix~\ref{app:imagenet}.

\begin{table}[h]
\centering
\caption{ImageNet $256{\times}256$ controlled comparison. Same encoder/decoder/AR architectures and training budget; wAR-Tok adds only the WGF-DPD term. Lower is better for all metrics.}
\label{tab:imagenet256}
\small
\begin{tabular}{lccc}
\toprule
Tokenizer & rFID & gFID & Eval AR Loss\\
\midrule
1D Tokenizer (baseline) & \textbf{2.05} & 6.32 & 7.89\\
\textbf{wAR-Tok (Ours)}    & 2.35 & \textbf{5.42} & \textbf{7.07}\\
\bottomrule
\end{tabular}
\end{table}

\section{Discussion}\label{sec:6}

In an idealized lossless bijection where $x\equiv z_{1:n}$, reconstructing from a prefix is equivalent to predicting the missing suffix, i.e., $\log p(x\mid z_{\le k})=\sum_{j=k+1}^{n}\log p(z_j\mid z_{<j})$. Tail dropout therefore resembles AR training in this special case. In practice, the equivalence is only approximate: the target is decoder reconstruction rather than prior likelihood, later tokens receive larger weight because they appear in more suffixes, and real tokenizers are not lossless bijections. SelfTok~\citep{selftok} relaxes the bijection assumption by tying prefix length to diffusion timestep and supervising tokens through a flow-matching decoder. Both methods nevertheless remain decoder-centric: they shape tokens through reconstruction or denoising rather than through the AR prior's own modeling loss. WGF-DPD targets this remaining mismatch by applying the AR prior's score directly to the encoder.

We deliberately exclude auxiliary techniques common in recent tokenizers (DINO alignment, distillation, diffusion-decoder post-training) to isolate the WGF-DPD effect, and therefore do not claim absolute SOTA. Following standard practice in image-tokenizer and AR-generation literature, where multi-seed error bars are uncommon due to per-run cost, we report single-seed numbers. Integration with the auxiliary techniques above and full-scale ablations on ImageNet are valuable next steps. We restrict attention to AR priors. Extending WGF-DPD to other discrete priors (MIM, masked diffusion) and exploring Cases~2 and~3 of TVC's redundancy are left to future work.

\bibliography{wgf_refs}

\appendix
\setcounter{tocdepth}{2}
\allowdisplaybreaks
\newpage

\section{Impact Statement}\label{app:impact}
\phantomsection\label{sec:impact}

Reducing the encoder--prior mismatch shrinks the parameter budget needed to fit a competitive AR prior, lowering the cost of training and serving AR-based generative models. We do not anticipate qualitatively new misuse risks beyond those already present in discrete generative models. Standard governance, content filtering, and ethical-oversight practices apply.

\section{Additional Experimental Results}\label{app:matched-iters}
Table~\ref{tab:compare_iters} complements the main matched-rFID comparison by controlling for total tokenizer training budget. Specifically, all methods are trained for the same number of iterations, which tests whether the gain of wAR-Tok comes from the proposed alignment objective rather than from longer optimization. Under this matched-iteration protocol, wAR-Tok keeps reconstruction quality comparable while substantially improving AR prior fit and generation quality.
\begin{table}[h]
\centering
\caption{CIFAR-10, comparison under \textbf{identical training budget (Iters $=$ 400K)}. With the same iterations, wAR-Tok achieves comparable reconstruction (rFID) and superior generation (gFID) quality.}
\label{tab:compare_iters}
\small
\begin{adjustbox}{width=\textwidth}
\begin{tabular}{lcccccccc}
\toprule
\multirow{2}{*}{Representative Method} & \multirow{2}{*}{Iters} & \multicolumn{4}{c}{Training Techniques} & \multirow{2}{*}{rFID $(\downarrow)$} & \multirow{2}{*}{gFID $(\downarrow)$} & \multirow{2}{*}{gLoss $(\downarrow)$}\\
\cmidrule(lr){3-6}
& & FM Dec. & Tail Drop. & Tied Steps & WGF-DPD & & & \\
\midrule
TiTok~\citep{titok}                       & 400K & \xmark & \xmark & \xmark & \xmark & 2.30 & 52.53 & 1.38\\
ImageFolder/One-D-Piece~\citep{imagefolder,onedpiece} & 400K & \xmark & \cmark & \xmark & \xmark & 3.35 & 23.37 & 0.93\\
FlowMo~\citep{sargent2025flowmodemodeseekingdiffusion}                     & 400K & \cmark & \xmark & \xmark & \xmark & 3.41 & 38.63 & 1.52\\
FlexTok~\citep{flextok}                   & 400K & \cmark & \cmark & \xmark & \xmark & 4.42 & 22.76 & 1.09\\
SelfTok~\citep{selftok}                   & 400K & \cmark & \cmark & \cmark & \xmark & 6.49 & 44.79 & 1.37\\
\midrule
\textbf{wAR-Tok (Ours)}                      & 400K & \xmark & \xmark & \xmark & \cmark & 3.06 & \textbf{11.30} & \textbf{0.34}\\
\bottomrule
\end{tabular}
\end{adjustbox}
\end{table}

\section{Related Work}\label{app:related}

\subsection{Comparison with Continuous-Space Distribution-Level Matching}\label{app:wae-comparison}
Table~\ref{tab:wae-info-comparison} summarizes why existing distribution-level matching methods do not directly solve the setting considered in this work. These methods share our high-level goal of matching the aggregate posterior with a prior, but they are designed for continuous latent spaces and rely on gradients through kernels, discriminators, or reparameterized samples. In contrast, our setting uses hard discrete tokenizers and an autoregressive prior, where such gradient routes either degenerate or become unstable. WGF-DPD provides a discrete-space alternative by using only forward AR log-probabilities to define the encoder update.
\begin{table}[h]
\centering
\caption{Comparison with distribution-level aggregate-posterior matching methods. On hard discrete quantizers the continuous-space methods either degenerate, become unstable, or require an intractable score, and WGF-DPD avoids all three. ``$\nabla_z\log q$'' denotes the analytic aggregate-posterior score.}
\label{tab:wae-info-comparison}
\small
\begin{adjustbox}{width=\textwidth}
\begin{tabular}{lcccc}
\toprule
& \textbf{WAE-MMD}~\citep{tolstikhin2018wae} & \textbf{WAE-GAN/AAE}~\citep{makhzani2015adversarial} & \textbf{InfoVAE}~\citep{zhao2017infovae} & \textbf{WGF-DPD (Ours)}\\
\midrule
Latent space & Continuous & Continuous & Continuous & \textbf{Discrete}\\
Objective & MMD & Adversarial & KL & KL\\
Gradient route & Backprop (kernel) & Backprop (discriminator) & Reparameterization & \textbf{Forward-pass via WGF}\\
Hard quantizer & Degenerate (kernel on one-hot) & Unstable (discrete GAN) & Intractable (no analytic $\nabla_z\log q$) & \textbf{Natively supported}\\
\bottomrule
\end{tabular}
\end{adjustbox}
\end{table}

\subsection{Further Related Work}

\noindent\textbf{Autoregressive modeling and large language models.} AR modeling has been the dominant paradigm for large-scale sequence generation. Transformer architectures~\citep{vaswani2017attention} enabled substantial scaling of next-token prediction. Extending AR to images dates back to PixelCNN~\citep{van2016pixelcnn}. Recent systems push AR generation to text-to-image synthesis~\citep{ramesh2021dalle} and LLM-style backbones~\citep{sun2024llamagen}.

\noindent\textbf{Discrete image generation.} Discrete LVMs commonly follow a two-stage recipe: learn a tokenizer, then fit a prior over discrete codes~\citep{esser2021taming}. Masked token modeling provides an alternative to left-to-right decoding~\citep{chang2022maskgit}.

\noindent\textbf{Wasserstein distance, gradient flow, and WAE.} Optimal transport and Wasserstein distances provide geometry-aware discrepancies between distributions~\citep{peyre2019computationalot}. Wasserstein gradient flows formalize steepest descent over probability measures~\citep{ambrosio2005gradient}. In generative modeling, Wasserstein objectives appear in WGAN~\citep{arjovsky2017wgan} and Wasserstein autoencoders~\citep{tolstikhin2018wae}, and motivate particle methods like SVGD~\citep{liu2016svgd}.

\noindent\textbf{Learnable priors and joint training.} In LVMs, prior design crucially impacts representation learning and downstream generation. Standard variational objectives~\citep{kingma2014vae} encourage posterior--prior matching but can suffer from posterior collapse~\citep{bowman2016generating}. This motivates learnable priors: VampPrior~\citep{tomczak2018vampprior}, normalizing-flow priors~\citep{rezende2015flows}, distribution-level aggregate-posterior matching~\citep{tolstikhin2018wae}, latent EBMs~\citep{du2019implicitebm}, and divergence-triangle joint training~\citep{han2019divergencetriangle}. Recent work emphasizes designing latents with downstream model complexity in mind~\citep{hu2023complexity}.

\noindent\textbf{1D tokenization and AR-alignment techniques.} Beyond rasterized patches, several tokenizers use learnable holistic queries~\citep{titok}. Tail-dropout-style objectives~\citep{flextok} implicitly induce causal structure. Diffusion/flow-based decoders~\citep{sargent2025flowmodemodeseekingdiffusion} introduce sequential structure through iterative refinement. SelfTok~\citep{selftok} couples token prefix length with diffusion timesteps. Some works~\citep{wang2024larp,ramanujan2024worsebetter} backpropagate the generative loss directly through the prior. Doing so picks up the maximum-likelihood term of the ELBO without the entropy term $H(q_\phi(z\mid x))$, and the encoder is pushed toward trivial sequences unless additional stabilizers and full backpropagation through the prior are added.

\section{Proofs}\label{sec:proof}
\subsection{Proof of Other Two Conditions in TVC}
\paragraph{Case 1 (Likelihood + Prior $\Rightarrow$ Posterior).}\label{Proof:TVC_case1}
The proof for Case 1 follows directly from the definition of the joint distribution.
\paragraph{Case 2 (Likelihood + Posterior $\Rightarrow$ Prior).}\label{Proof:TVC_case2}
Assume $p_\theta(z\mid x)=q_\phi(z\mid x)$ and $p_\theta(x\mid z)=q_\phi(x\mid z)$. Then
\begin{align*}
p_\theta(z)
&= \frac{p_\theta(z\mid x)p_\theta(x)}{p_\theta(x\mid z)} \nonumber \\
&= \frac{q_\phi(z\mid x)p_\theta(x)}{q_\phi(x\mid z)}
 = \frac{p_\theta(x)}{\pdata(x)}\,q_\phi(z).
\end{align*}
Since the left-hand side does not depend on $x$, the ratio $p_\theta(x)/\pdata(x)$ must be constant in $x$. Integrating both sides over $z$ yields $p_\theta(x)=\pdata(x)$ and hence $p_\theta(z)=q_\phi(z)$.
\paragraph{Case 3 (Posterior + Prior $\Rightarrow$ Likelihood under completeness).}
Assume $p_\theta(z\mid x)=q_\phi(z\mid x)$ and $p_\theta(z)=q_\phi(z)$. Then
\begin{align*}
p_\theta(x\mid z)
&= \frac{p_\theta(z\mid x)p_\theta(x)}{p_\theta(z)}
 = \frac{q_\phi(z\mid x)p_\theta(x)}{q_\phi(z)} \nonumber \\
&= q_\phi(x\mid z)\,\frac{p_\theta(x)}{\pdata(x)}.
\end{align*}
Integrating both sides over $x$ gives
\begin{equation*}
\E_{x\sim q_\phi(x\mid z)}\!\left[\frac{p_\theta(x)}{\pdata(x)}\right] = 1,\quad \forall z.
\end{equation*}
If the family $\{q_\phi(x\mid z)\}_z$ is complete, the above implies $p_\theta(x)=\pdata(x)$ (a.e.), and substituting back yields $p_\theta(x\mid z)=q_\phi(x\mid z)$.

\subsection{Proof of ELBO as an Instance of TVC}\label{Proof:ELBO_TVC}
Using MLE to optimize the log likelihood, which equals to minimize the data-model marginal distribution with KL, first expand to TVC form, and it can be clearly convert to ELBO:
\begin{equation*}
\begin{aligned}
   &\E_{x\sim \pdata}\big[\log p_\theta(x)\big] \\
    &= \E_{q_\phi(x,z)}\Bigg[
        \log \frac{p_\theta(x\mid z)}{q_\phi(x\mid z)}
      + \log \frac{p_\theta(z)}{q_\phi(z)} 
       - \log \frac{p_\theta(z\mid x)}{q_\phi(z\mid x)}
     + \log \pdata(x) \Bigg]  \\
    &= \E_{q_\phi(x,z)}\Bigg[
        \log p_\theta(x\mid z) - \log \frac{q_\phi(z\mid x) \pdata(x)}{q_\phi(z)} \\
      &\quad\quad\quad + \log \frac{p_\theta(z)}{q_\phi(z)}
      - \log \frac{p_\theta(z\mid x)}{q_\phi(z\mid x)}
     + \log \pdata(x) \Bigg] \\
    &= \underbrace{\E_{x\sim \pdata}\Big[\E_{z\sim q_\phi(z\mid x)}\big[\log p_\theta(x\mid z)\big]
      - \KL\!\left(q_\phi(z\mid x)\,\|\,p_\theta(z)\right)\Big]}_{\text{ELBO}}  \\
    &\quad + \underbrace{\E_{x\sim \pdata}\Big[\KL\!\left(q_\phi(z\mid x)\,\|\,p_\theta(z\mid x)\right)\Big]}_{\text{Posterior Consistency Gap}}.
    \label{eq:elbo_decomp}
\end{aligned}
\end{equation*}

This derivation demonstrates that the ELBO is fundamentally a specific instantiation of the TVC formalism, which targets Likelihood Consistency and Prior Consistency via KL divergence. The residual term corresponds to the well-known variational gap, which vanishes when the bound is tight. Consistent with our Constraint Redundancy analysis, this perspective confirms that explicitly satisfying likelihood and prior consistency provides a sufficient guarantee for posterior consistency.

\subsection{Proof of ELBO Variants as an Instance of TVC}\label{Proof:ELBOVariants_TVC}
By the redundancy property (Corollary~\ref{lem:redundancy}, Case~1), global optimality $p_\theta(x)=\pdata(x)$ can be achieved by enforcing Likelihood Consistency and Prior Consistency. The standard ELBO enforces prior consistency through an instance-wise KL term $\E_{x\sim \pdata}\KL(q_\phi(z\mid x)\|p_\theta(z))$, which is sufficient but can be overly restrictive and induce posterior collapse.

A broad class of ``ELBO variants'' can be interpreted as replacing this instance-wise prior matching by a \emph{distribution-level} penalty between marginals, optionally coupled with information/structure regularizers on the aggregate posterior:
\begin{equation*}
 \max_{\theta,\phi}\;
 \E_{x\sim \pdata}\E_{z\sim q_\phi(z\mid x)}\!\big[\log p_\theta(x\mid z)\big]
 \;-\;\lambda\,\mathcal{D}\!\left(q_\phi(z),\,p_\theta(z)\right)
 \;+\;\gamma\, I_{q_\phi}(X;Z)
 \;-\;\eta\,\mathcal{R}\!\left(q_\phi(z)\right),
\end{equation*}
where $q_\phi(z)\triangleq \E_{\pdata}[q_\phi(z\mid x)]$, $\mathcal{D}$ is a divergence between \emph{marginals} (e.g., adversarial $f$-divergences, MMD, or Wasserstein), $I_{q_\phi}(X;Z)$ discourages the degenerate solution $z\!\perp\! x$, and $\mathcal{R}$ optionally regularizes aggregate structure (e.g., total correlation).
Whenever optimizing such an objective yields (approximately) Likelihood Consistency $p_\theta(x\mid z)\approx q_\phi(x\mid z)$ and Prior Consistency $p_\theta(z)\approx q_\phi(z)$, the model enters the Case~1 consistency regime and therefore drives the same marginal objective $p_\theta(x)\approx \pdata(x)$.

\subsection{Proof of Two-Stage Training as an Instance of TVC}\label{Proof:TwoStage_TVC}

The training protocol explicitly targets two TVC conditions. In the first stage, the encoder and decoder are optimized to maximize the reconstruction likelihood $\E_{z\sim q_\phi(z\mid x)}[\log p_\theta(x\mid z)]$, which achieves its global optimum if and only if Likelihood Consistency holds: $p_\theta(x\mid z) = q_\phi(x\mid z)$. In the second stage, the encoder is frozen, and a prior model $p_\theta(z)$ (e.g., autoregressive) is trained to minimize $\KL(q_\phi(z)\,\|\,p_\theta(z))$, where $q_\phi(z) \triangleq \E_{\pdata}[q_\phi(z\mid x)]$. Convergence of this stage ensures Prior Consistency: $p_\theta(z) = q_\phi(z)$. By Corollary \ref{lem:redundancy}, the simultaneous satisfaction of these two conditions is sufficient to imply Posterior Consistency and guarantee $p_\theta(x) = \pdata(x)$, thereby providing a rigorous theoretical justification for the two-stage paradigm.

\subsection{Proof of TVC}\label{Proof:TVC_main}
\begin{equation*}
    \centering
    \begin{aligned}
    \log p_\theta(x)
    &= \log \frac{p_\theta(x\mid z)p_\theta(z)}{p_\theta(z\mid x)} \\
    &= \log \frac{p_\theta(x\mid z)p_\theta(z)}{p_\theta(z\mid x)}
     + \log \frac{q_\phi(x,z)}{q_\phi(x,z)} \\
    &= \log \frac{p_\theta(x\mid z)p_\theta(z)}{p_\theta(z\mid x)}
     + \log \frac{q_\phi(z\mid x)\pdata(x)}{q_\phi(x\mid z)q_\phi(z)} \\
    &= \underbrace{\log \frac{p_\theta(x\mid z)}{q_\phi(x\mid z)}}_{\text{Likelihood Consistency}}
     + \underbrace{\log \frac{p_\theta(z)}{q_\phi(z)}}_{\text{Prior Consistency}} \\
     &\quad- \underbrace{\log \frac{p_\theta(z\mid x)}{q_\phi(z\mid x)}}_{\text{Posterior Consistency}}
     + \log \pdata(x)
    \end{aligned}
\end{equation*}
\section{Wasserstein Gradient Flow}\label{chap:AppendixB}

Under the Wasserstein-2 distance \citep{peyre2019computationalot}, we define
\begin{equation*}
W_2^2(Q, P) = \inf_{\gamma \in \Gamma(Q, P)} \int_{X \times X} \|x - y\|^2 \, \mathrm{d}\gamma(x, y).
\end{equation*}

Introduce the continuity equation \citep{Otto31012001}:
\begin{equation*}
\frac{\partial \rho_t}{\partial t} + \nabla \cdot (\rho_t v_t) = 0.
\end{equation*}

The Wasserstein-2 distance admits an equivalent dynamic (minimum kinetic energy) formulation \citep{benamou2000computational}:
\begin{equation*}
W_2^2(\rho_0, \rho_1) =
\inf_{\rho_t, v_t}
\left\{
\int_0^1 \int_X \|v_t(x)\|^2 \rho_t(x)\, \mathrm{d}x \,\mathrm{d}t
\right\}
\end{equation*}
subject to $\frac{\partial \rho_t}{\partial t} + \nabla \cdot (\rho_t v_t) = 0$.

In this setting, the time derivative of an energy functional $F[\rho]$ is
\begin{align*}
\frac{\mathrm{d} F[\rho_t]}{\mathrm{d} t}
&=
\int_X \frac{\delta F}{\delta \rho_t}(x)\, \frac{\partial \rho_t}{\partial t}(x)\, \mathrm{d}x \nonumber \\
&=
\int_X \frac{\delta F}{\delta \rho_t}(x)
\left( - \nabla \cdot (\rho_t(x) v_t(x)) \right)\, \mathrm{d}x \nonumber \\
&=
\int_X \left( \nabla \frac{\delta F}{\delta \rho_t}(x) \right)\cdot v_t(x)\, \rho_t(x)\, \mathrm{d}x.
\end{align*}

Meanwhile, the Wasserstein gradient $V$ is defined via
\begin{equation*}
\frac{\mathrm{d} F[\rho_t]}{\mathrm{d} t}
=
\langle V, v_t \rangle_{L^2(\rho_t)}
=
\int_X V(x)\cdot v_t(x)\, \rho_t(x)\, \mathrm{d}x.
\end{equation*}
Solving for $V$ gives
\begin{equation*}
V(x) = \nabla \left( \frac{\delta F}{\delta \rho_t}(x) \right).
\end{equation*}

The Wasserstein gradient flow defined a the gradient of a functional to samples as:
\begin{equation*}
v_t(x) = -V(x) = - \nabla \left( \frac{\delta F}{\delta \rho_t}(x) \right).
\end{equation*}

\section{Implementation Details and Architectures}\label{chap:AppendixC}

\subsection{CIFAR-10 Setup}

\noindent\textbf{Architecture.} We adopt a modern Transformer backbone across all components, incorporating GEGLU, RMSNorm, and AdaRMSNorm. Table~\ref{tab:arch} summarizes the dimensions used on CIFAR-10.

\begin{table}[h]
\centering
\caption{Architectural details used in our CIFAR-10 experiments.}
\label{tab:arch}
\small
\begin{tabular}{lccccc}
\toprule
Component & Layers & Dim & Heads & MLP Ratio & Params\\
\midrule
Encoder                       & 10 & 512 & 8 & 4 & $\sim$50M\\
Decoder                       & 10 & 512 & 8 & 4 & $\sim$50M\\
AR proxy $q_\psi$ (joint)     &  4 & 512 & 8 & 4 & $\sim$17M\\
AR target prior $p_\theta(z)$ & 13 & 512 & 8 & 4 & $\sim$50M\\
\bottomrule
\end{tabular}
\end{table}

\noindent\textbf{Training.} For Stage-I training on CIFAR-10, we train encoder and decoder for 400K iterations at batch size 512. For SelfTok we extend Stage-I by an additional 100K iterations to reach comparable rFID. For deterministic decoders we use weighted $\ell_1$ ($0.1$), $\ell_2$ ($1.0$), and LPIPS~\citep{zhang2018lpips} ($0.1$). For diffusion decoders, following FlowMo~\citep{sargent2025flowmodemodeseekingdiffusion}, we use a flow-matching loss and compute LPIPS on $\hat x_0$ converted from $v$-prediction. For AR training on CIFAR-10, we use a 50M-parameter AR prior trained for a fixed 400K iterations. We use L2-normalized vector quantization (VQ) as the default quantizer.

We implement tail dropout following SelfTok: sample $t\sim\mathcal{U}(0,1)$ and map to a discrete cut-off index using a SelfTok-like scheduler that biases toward later token positions. For non-tied-timestep methods we enable tail dropout from the start with probability $0.5$. For tied-timestep methods the ratio is determined by diffusion timesteps and tail dropout is applied at every iteration. We keep diffusion and deterministic decoders architecturally identical except for input format and timestep conditioning through AdaRMSNorm.

\noindent\textbf{WGF-DPD-specific details.} For our WGF-DPD training, we map L2-normalized Euclidean distances between latent $z$ and codebook entries to logits and construct one-hot assignments using a probability-STE adapted from the hard Gumbel-Softmax STE. We use a lightweight AR proxy $q_\psi$ as the source distribution. AR loss, WGF-gradient term, and reconstruction loss are optimized jointly in a single end-to-end run (no alternating updates), incurring about 10\% additional cost over reconstruction-only training. Since the tokenizer-induced aggregate posterior is generally not autoregressive-friendly early on, using a target AR model $p_\theta$ that is trained to fit this non-AR distribution would weaken the incentive for the tokenizer to produce AR-friendly latents (the target can adapt instead). We therefore fix the target to a randomly-initialized AR Transformer $p_{\theta_0}$ (with a causal mask) and apply a low temperature to sharpen its predictive distribution. At initialization, attention weights are approximately uniform within the causal context across local and global positions. Lowering the temperature increases the target's confidence and amplifies this AR-biased guidance signal during joint training.

\noindent\textbf{Stability.} No heuristic stabilizers are used. All losses are optimized jointly with standard AdamW. We observe no instability across runs.

\noindent\textbf{Compute.} CIFAR-10 experiments run on $8\times$ NVIDIA RTX 5880 GPUs. Tokenizer training (about 400K iters, batch size 512) takes approximately one day. AR training (400K iters, $\sim$50M parameters) takes approximately half a day.

\subsection{ImageNet 256 Setup}\label{app:imagenet}

\noindent\textbf{Losses.} $\ell_1$ (weight $1.0$), LPIPS (weight $1.0$), PatchGAN (weight $0.1$).

\noindent\textbf{Model sizes.} Encoder/decoder $\sim$86M total (tokenizer); AR target prior $\sim$115M; AR proxy $\sim$40M for joint training.

\noindent\textbf{Optimizer.} AdamW with weight decay $3\times10^{-2}$ for both the tokenizer (including the joint-training AR proxy) and the AR target prior.

\noindent\textbf{Training.} 150 epochs (tokenizer), 300 epochs (AR).

\noindent\textbf{Compute.} $8\times$ NVIDIA A100 GPUs. Tokenizer training takes about 30 hours, AR training about 20 hours.

\noindent\textbf{Evaluation.} FID-50K on the ImageNet validation set. We search the CFG schedule per method. The baseline uses a cosine schedule with scale 20 and power 1.5, and wAR-Tok uses a cosine schedule with scale 24 and power 2.0. The baseline is a standard 1D Tokenizer without auxiliary tricks. Due to compute constraints, hyperparameters are not finely tuned, and the table reports relative improvement over the controlled baseline rather than absolute SOTA. rFID is computed on the validation set, and gFID follows the ADM protocol. We observe wAR-Tok also achieves faster gFID convergence than the baseline throughout training.

\subsection{Why we focus on AR priors here}\label{app:why_ar}
TVC and DPD are prior-agnostic. WGF-DPD applies to any discrete prior with tractable log-probabilities. We focus on AR priors because they dominate language modeling yet remain underexplored in visual/multimodal generation, and because the data--model prior mismatch we target is most acute here.

\section{Qualitative Samples}\label{app:samples}

\begin{figure}[t]
\centering
\begin{subfigure}[b]{0.16\linewidth}\centering\includegraphics[width=\linewidth]{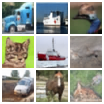}\caption*{\centering Ours\\gFID=11.30}\end{subfigure}\hfill
\begin{subfigure}[b]{0.16\linewidth}\centering\includegraphics[width=\linewidth]{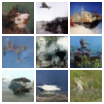}\caption*{\centering Baseline\\gFID=52.53}\end{subfigure}\hfill
\begin{subfigure}[b]{0.16\linewidth}\centering\includegraphics[width=\linewidth]{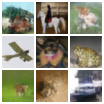}\caption*{\centering Tail.\\gFID=23.37}\end{subfigure}\hfill
\begin{subfigure}[b]{0.16\linewidth}\centering\includegraphics[width=\linewidth]{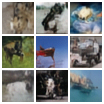}\caption*{\centering Diff.\\gFID=38.63}\end{subfigure}\hfill
\begin{subfigure}[b]{0.16\linewidth}\centering\includegraphics[width=\linewidth]{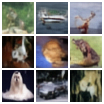}\caption*{\centering Diff.+Tail.\\gFID=22.76}\end{subfigure}\hfill
\begin{subfigure}[b]{0.16\linewidth}\centering\includegraphics[width=\linewidth]{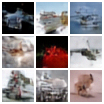}\caption*{\centering Diff.+Tail.+Tied.\\gFID=44.79}\end{subfigure}
\caption{Qualitative comparison on CIFAR-10. wAR-Tok produces samples with substantially better visual quality and the lowest gFID. ``Diff.'' = Diffusion Decoder; ``Tail.'' = Tail Dropout; ``Tied.'' = Tied Timesteps.}
\label{fig:visual_comparison}
\end{figure}

We provide qualitative results in Figure~\ref{fig:visual_comparison}. For a fair comparison, we use fixed random seeds across methods to control for sampling variability.

\end{document}